\documentclass[lettersize,journal]{IEEEtran}
\usepackage{amsmath,amsfonts}
\usepackage{algorithm}
\usepackage{array}
\usepackage[caption=false,font=normalsize,labelfont=sf,textfont=sf]{subfig}

\usepackage{verbatim}
\usepackage{cite}
\usepackage{multirow}
\usepackage{multicol}

\usepackage{graphicx}
\usepackage{booktabs}

\usepackage{times}
\usepackage{epsfig}
\usepackage{diagbox}
\usepackage{graphics}
\usepackage{threeparttable}
\usepackage{xspace}
\usepackage{siunitx}
\usepackage{color}
\usepackage[normalem]{ulem}
\usepackage{multirow}
\usepackage{float}
\usepackage{amsfonts}
\usepackage{bm}
\usepackage[table]{xcolor}
\usepackage{colortbl}
\usepackage{diagbox}
\usepackage{rotating}
\usepackage{booktabs}
\usepackage{overpic}
\usepackage{enumitem}
\usepackage{pifont}
\usepackage{dblfloatfix}
\usepackage{wrapfig,lipsum,booktabs}
\usepackage[utf8]{inputenc}
\usepackage[pagebackref,breaklinks,colorlinks]{hyperref}
\definecolor{bblue}{rgb}{0,150,230}
\definecolor{mygray}{gray}{.9}
\definecolor{myy}{RGB}{126,95,0}

\makeatletter
\newcommand{\thickhline}{%
	\noalign {\ifnum 0=`}\fi \hrule height 1pt
	\futurelet \reserved@a \@xhline
}
\makeatother

\makeatletter
\DeclareRobustCommand\onedot{\futurelet\@let@token\@onedot}
\def\@onedot{\ifx\@let@token.\else.\null\fi\xspace}

\makeatother

\usepackage{cleveref}
\crefname{section}{§}{§§}

\usepackage[T1]{fontenc}    
\usepackage{url}            
\usepackage{nicefrac}       
\usepackage{microtype}      
\begin{document}
	
	\title{Rethinking the Encoding and Annotating of 3D Bounding Box: \\ Corner-Aware 3D Object Detection from Point Clouds}
	
\author{
    Qinghao Meng,
    Junbo Yin,
    Jianbing Shen,
    Yunde Jia

\thanks{ Q. Meng is with the School of Computer Science, Beijing Institute of Technology, Beijing, China. {\tt\small mengqinghao@bit.edu.cn}}
\thanks{ J. Yin is with Computer Science Program, Computer, Electrical and Mathematical Sciences and Engineering (CEMSE) Division, Center of Excellence for Smart Health, and Center of Excellence for Generative AI, King Abdullah University of Science and Technology (KAUST), Thuwal 23955-6900, Kingdom of Saudi Arabia. {\tt\small junbo.yin@kaust.edu.sa}}
\thanks{ J. Shen is with the State Key Laboratory of Internet of Things for Smart City, Department of Computer and Information Science, University of Macau, Macau, China. {\tt\small jianbingshen@um.edu.mo}}
\thanks{ Y. Jia is with the Guangdong Provincial Key Laboratory of Machine Perception and Intelligent Computing, Shenzhen MSU-BIT University, China, and with the Beijing Key Laboratory of Intelligent Information Technology, School of Computer Science, Beijing Institute of Technology, China. {\tt\small jiayunde@smbu.edu.cn}}

\thanks{ Corresponding author: \textit{J. Shen} and \textit{Y. Jia}.}

}

\maketitle

\begin{abstract}
		
Center-aligned regression remains dominant in LiDAR-based 3D object detection, yet it suffers from fundamental instability: object centers often fall in sparse or empty regions of the bird's-eye-view (BEV) due to the front-surface-biased nature of LiDAR point clouds, leading to noisy and inaccurate bounding box predictions. To circumvent this limitation, we revisit bounding box representation and propose corner-aligned regression, which shifts the prediction target from unstable centers to geometrically informative corners that reside in dense, observable regions. Leveraging the inherent geometric constraints among corners and image 2D boxes, partial parameters of 3D bounding boxes  can be recovered from corner annotations, enabling a weakly supervised paradigm without requiring complete 3D labels. We design a simple yet effective corner-aware detection head that can be plugged into existing detectors. Experiments on KITTI show our method improves performance by 3.5\% AP over center-based baseline, and achieves 83\% of fully supervised accuracy using only BEV corner clicks, demonstrating the effectiveness of our  corner-aware regression strategy.
		
	\end{abstract}

\section{Introduction}
\label{sec:intro}

LiDAR-based 3D object detection has been largely driven by deep learning architectures adapted from 2D vision~\cite{Shi_2019_CVPR,Lang_2019_CVPR,shi2020pv}. A dominant design choice across most methods is the use of \textit{center-aligned} bounding box regression, where a 3D object is represented by seven parameters: its center coordinates $(x, y, z)$, dimensions $(l, w, h)$, and yaw angle $\theta$. This formulation inherits the simplicity and efficiency of 2D center-based detectors such as CenterNet~\cite{duan2019centernet}, enabling end-to-end training with relatively straightforward supervision. However, this paradigm implicitly assumes that the object center is a well-supported and informative anchor for regression—a premise that in dense image domains but breaks down in sparse 3D point clouds.

\begin{figure*}[t]
		\vspace{-4mm}
		\centering
		\includegraphics[width=0.94\linewidth]{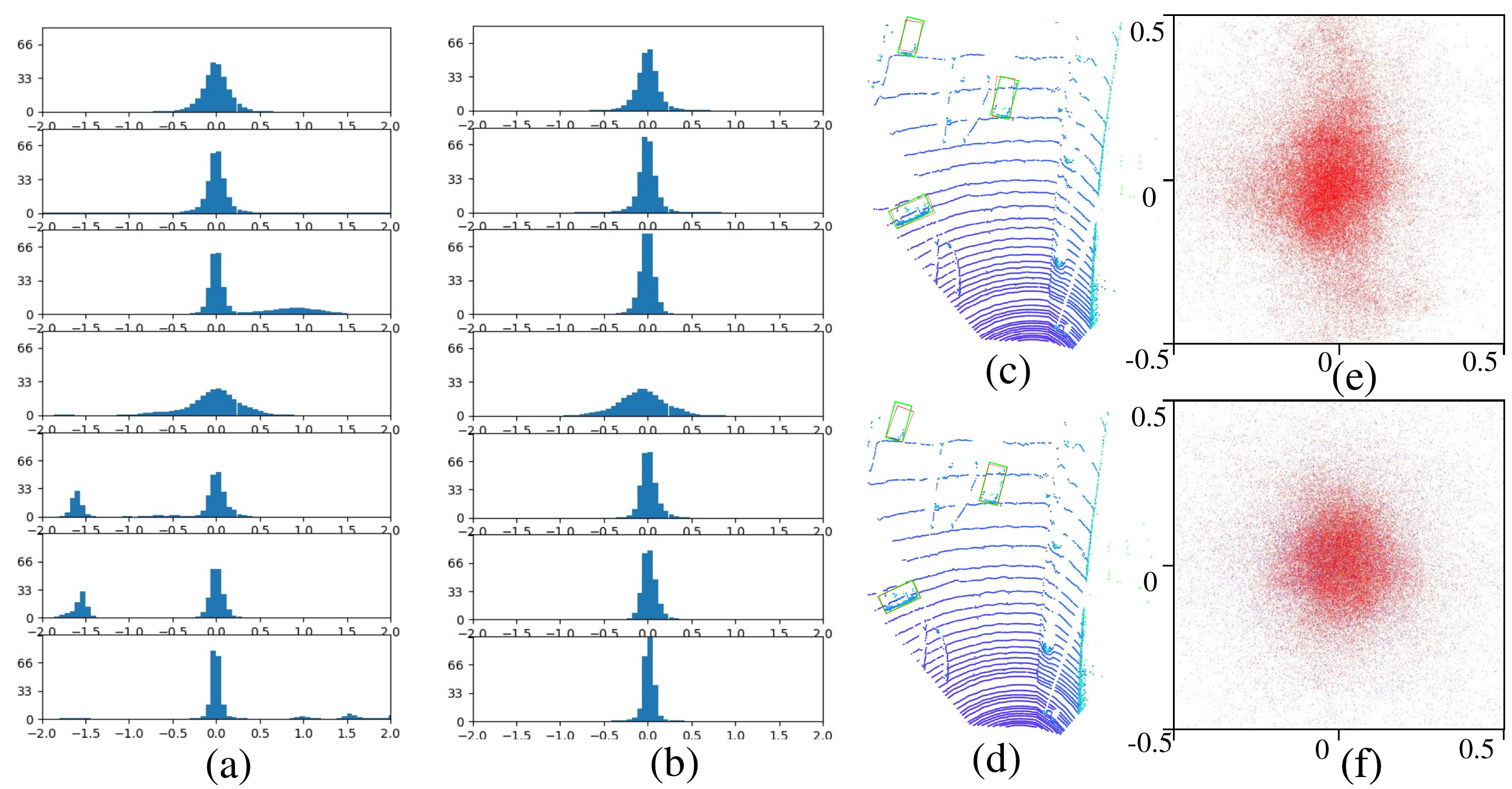}
		\caption{Comparisons of 3D bounding box proposal prediction errors (x, y, z, length, width, height, yaw-angle) between baseline ( PointRCNN~\!~\cite{Shi_2019_CVPR}) and the proposed method \textbf{(a-b)}, the visualization of point cloud scene between baseline and the one with our improvement \textbf{(c-d)},
		and the comparisons of location error \textbf{(e-f)}.
		}
		\label{fig:statictics}
		\vspace{-5mm}
	\end{figure*}

The core issue stems from the intrinsic sensing characteristics of LiDAR. Due to occlusion, limited sensor range, and the physics of laser reflection, LiDAR point clouds are highly non-uniform and exhibit a strong \textit{front-surface bias}: points are densely distributed on the visible front and sides of vehicles, while rear surfaces, tops, and interior regions often remain unobserved. As a result, the geometric center of an object frequently lies in empty or extremely sparse regions in the bird’s-eye-view (BEV) plane. When a neural network regresses a full 3D bounding box from such a poorly grounded center, it must extrapolate critical geometric attributes—such as orientation and size—using only indirect surface-level features. This leads to high variance in predictions: even minor errors in yaw or dimension estimation can cause large displacements of box corners, dramatically reducing intersection-over-union (IoU) despite reasonably accurate center localization~\cite{ye2020improving}. Moreover, this misalignment between classification confidence and actual localization quality complicates post-processing (e.g., non-maximum suppression) and caps detection performance. As illustrated in Fig.~\ref{fig:statictics}, corner-aligned regression yields tighter error distributions across all seven box dimensions, particularly for height, width, and yaw, and achieves superior geometric consistency in final predictions.

The corner-aligned regression strategy not only enhances robustness under full supervision when integrated into existing detectors, but also facilitates weakly supervised learning. We introduce a practical annotation protocol that requires only \textit{clicks on visible BEV corners}, which are far easier and faster to label than full 3D bounding boxes. By leveraging geometric relationships among annotated corners and incorporating height priors from 2D image detections, we reconstruct approximate 3D bounding boxes as supervision signals. Our framework operates in two stages: (1) corner prediction and clustering to generate region proposals, and (2) 3D bounding box recovery using geometric constraints and cross-modal priors. Evaluated on the KITTI benchmark, our method achieves 83\% of fully supervised performance without any 3D bounding box annotations (using BEV corner annotations only), demonstrating its viability under extreme annotation scarcity. To support future research, we will publicly release our BEV corner annotations for KITTI.	
    Our contributions are summarized as follows:
\begin{itemize}
    \item We identify the inherent instability of center-aligned regression in sparse LiDAR scenes and advocate \textit{corner-aligned regression} as a geometrically principled alternative.
\item  Through experiments across multiple corner encoding schemes, we identify \textit{full-corner encoding} (F-Corner)  as the most effective representation, yielding the largest gain in average precision among all variants.
    \item We propose a practical BEV corner-click annotation scheme and a two-stage weakly supervised framework that recovers full 3D bounding boxes from partial corner labels and 2D height priors.
    \item We demonstrate that high-quality 3D detection is achievable without any 3D bounding box annotations, attaining 83\% of fully supervised accuracy on KITTI, and release our corner annotations to facilitate community research.
\end{itemize}

\section{Related Works}\label{sec:related_works}

\noindent\textbf{LiDAR-based 3D object detection.}
Most existing LiDAR-based 3D detectors adopt a \textit{center-aligned} regression framework, where the target is parameterized by the object center, size, and orientation~\cite{Shi_2019_CVPR,Lang_2019_CVPR,yang20203dssd}. These methods process point clouds either via BEV projection with 2D CNNs~\cite{Yang_2018_CVPR}, voxelization with 3D/sparse convolutions~\cite{Zhou_2018_CVPR,Yan2018}, or direct point-wise learning using PointNet-style backbones~\cite{Qi_2017_CVPR,Shi_2019_CVPR}. While effective, all these approaches share the same regression target: the geometric center. Some works introduce auxiliary losses involving corners or box sub-regions to enrich feature learning~\cite{Shi_2019_part,bhattacharyya2021self,Chen_2019}, but corners are never used as the \textit{primary regression objective}. In contrast, we propose to directly regress BEV corners as the main learning signal, leveraging their geometric consistency and alignment with dense point regions.

\noindent\textbf{Weakly supervised 3D detection and annotation reduction.}
Reducing the cost of 3D annotation has attracted growing attention. Early efforts use 2D seeds or templates to infer 3D proposals~\cite{zakharov2019autolabeling,lee2018leveraging}. Some researches explore weak supervision from sparse labels: WS3D~\textit{et al.}~\cite{meng2020weakly,meng2021towards} train detectors using only BEV center clicks, supplemented by a few fully annotated samples. Auto4d~\cite{yang2021auto4d} leverage temporal consistency in LiDAR sequences to generate pseudo-labels, and notably observe that corner-aligned representations yield lower noise than center-aligned ones—though their focus remains on tracking, not detection under pure weak supervision. Crucially, \textit{all existing weakly supervised methods still rely on center-based box encoding}.

\noindent\textbf{Geometric reasoning for 3D recovery.}
In monocular 3D detection, geometric constraints between 2D projections and 3D bounding boxes are widely used to recover depth and orientation~\cite{Chen_2016_CVPR,mousavian20173d,Li_2019_CVPR}. However, these methods suffer from inherent depth ambiguity. In our setting, we exploit a similar idea—but in a LiDAR-centric context: given BEV corner annotations (which lack height), we recover the missing $z$-dimension by leveraging height priors from 2D image detections and enforcing rigid box geometry. Unlike image-based methods that \textit{estimate} unreliable depth, we \textit{complete} a partially observed 3D structure using accurate LiDAR geometry and cross-modal height cues, enabling high-fidelity weak supervision.

This center-centric paradigm becomes particularly problematic under sparse point distributions, motivating our shift to corner-aligned representation
	
	\section{Method}

\subsection{Definition of Bounding Box Encoding Schemes}
To explore the optimal corner-aware 3D bounding box encoding, this paper systematically analyzes multiple corner-aware object representations. We first review the standard encoding scheme in 2D object detection and then generalize it to the 3D scenario.

\noindent\textbf{2D Object Detection.}  
In the anchor-based detection framework used by Zhao~\textit{et al.}~\cite{zhao2019object}, anchors are typically placed on image grid locations, generating multi-scale anchor boxes whose centers and parameters are refined via regression. A 2D bounding box $B_{2D}$ is defined as
\begin{equation}
    \begin{aligned}
        B_{2D} = \left( x_{c}, y_{c}, l, w \right),
    \end{aligned}
    \label{eq:b2d}
\end{equation}
where $(x_{c}, y_{c})$ denotes the pixel coordinates of the object center, and $(l, w)$ represents the box lengths along the $x$- and $y$-axes, respectively. This representation can be normalized into the interval $(-1, 1)$ to suit neural network training. Let the anchor be located at grid position $(x_a, y_a)$ with anchor dimensions $(l_a, w_a)$. The bounding box can then be parameterized via relative offsets as
\begin{equation}
    \begin{aligned}
        B_{2D} &= {B_{2D}}_{a} \times \left( \left[ m_x, m_y, 1, 1 \right] + {B_{2D}}_{r}[:4] \right), \\
        {B_{2D}}_{r} &= \left( x_{r}, y_{r}, l_{r}, w_{r}, s_{f} \right), ~~~~~~~~~~~~~~~~ \\
        {B_{2D}}_{a} &= \left( x_{a}, y_{a}, l_{a}, w_{a} \right), ~~~~~~~~~~~~~~~~~~~~~
    \end{aligned}	
    \label{eq:b2d_formulated}
\end{equation}
where $(m_x, m_y) \subseteq \mathbb{R}^2$ indicates the grid position of the anchor, $(x_{r}, y_{r})$ denotes the center offset, and $(l_{r}, w_{r})$ represents the scale factors. The term $s_{f} \in \{0, 1\}$ is the objectness confidence score (1 for positive samples). All regression targets are normalized into the range $(-1, 1)$ and serve as the network’s prediction target ${B_{2D}}_{p}$:
\begin{equation}
    \begin{aligned}
        {B_{2D}}_{p} = \left( x_{r}', y_{r}', l_{r}', w_{r}', s_{f}' \right), ~~~ \\
        \mathcal{L}_{2D} = \mathcal{L} \left( {B_{2D}}_{r} - {B_{2D}}_{p} \right).~~
    \end{aligned}	
    \label{eq:lb2d}
\end{equation}

\noindent\textbf{3D Object Detection.}  
Given a 3D point cloud $P = \{ p_i \}_{i=1}^N \subset \mathbb{R}^4$ containing $N$ points, where each point $p_i = (x_i, y_i, z_i, r_i)$ consists of 3D coordinates and reflectance intensity, 3D detection further requires estimating the object height $h$, vertical center $z$, and yaw angle $\theta$, beyond the 2D case. When extending the image grid to a voxel grid, the 3D bounding box can be analogously represented as
\begin{equation}
\begin{aligned}
    B_{3D} &= {B_{3D}}_{a} \times \bigl( [m_x, m_y, m_z, 1, 1, 1, m_{\theta}] + {B_{3D}}_{r}[:7] \bigr), \\
    {B_{3D}}_{a} &= \bigl( x_{a}, y_{a}, z_{a}, l_{a}, w_{a}, h_{a}, \theta_{a} \bigr), \\
    {B_{3D}}_{r} &= \bigl( x_{r}, y_{r}, z_{r}, l_{r}, w_{r}, h_{r}, \theta_{r}, s_{f} \bigr), \\
    {B_{3D}}_{p} &= \bigl( x_{r}', y_{r}', z_{r}', l_{r}', w_{r}', h_{r}', \theta_{r}', s_{f}' \bigr), \\
    \text{Loss}_{3D} &= \mathcal{L} \bigl( {B_{3D}}_{r} - {B_{3D}}_{p} \bigr),
\end{aligned}
\label{eq:b3d_formulate}
\end{equation}
where $m_u \cdot u_a$ ($u \in \{x, y, z, \theta\}$) denotes the index of the anchor in the voxel grid and the yaw angle bin. Some methods encode the angle using $\sin\theta$ and $\cos\theta$ to avoid periodic ambiguity, i.e., $\theta = \arctan2(\sin\theta, \cos\theta)$. For point-based methods (as opposed to voxelization), the anchor position can be replaced by the actual point coordinates $(x_i, y_i, z_i)$.

\begin{figure*}[h!]
	\centering
	\includegraphics[width=0.99\linewidth]{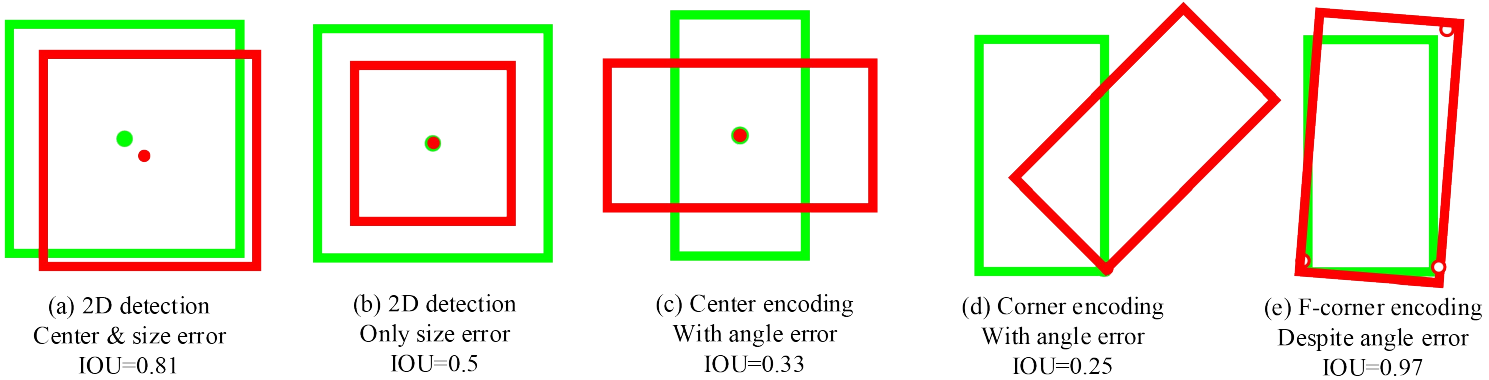}
	\caption{Impact of prediction errors on IoU for different object detection formulations.}
	\label{fig:iou}
\end{figure*}

While this center-based parameterization is widely adopted, it exhibits inherent sensitivity to regression errors in critical parameters—especially the yaw angle $\theta$ and vertical center $z$. As shown in Fig.~\ref{fig:iou}(c), even a small angular deviation can cause a dramatic drop in IoU, rendering the detection unreliable. Moreover, the object center often lies in sparsely populated or empty regions of the point cloud, making its precise localization challenging and annotation-intensive. These limitations motivate us to explore alternative representations that are more geometrically grounded and robust to perturbations.

In the following section, we shift our focus from center-based to corner-aware encoding schemes, leveraging the fact that corners are typically well-supported by dense LiDAR points and provide direct geometric constraints for box reconstruction.

\subsection{Corner Representation of 3D bounding boxes in Bird’s-Eye View}
\label{subs:corner_representation}

Building on the limitations of center-based encoding identified in above, we now investigate corner-aware representations
which effectively reduce the cumulative regression error at long distances. However, directly aligning to corner coordinates is sensitive to pose perturbations. Therefore, this paper explores corner representations that are robust to parameter disturbances and compares the strengths and weaknesses of five encoding strategies.

As illustrated in Fig.~\ref{fig:iou}(a–b), local parameter errors have relatively limited impact on IoU in 2D detection. However, in 3D detection, the additional sensitive parameters—particularly $z$, $h$, and $\theta$—make the task significantly more vulnerable. Even a small yaw error can drastically reduce IoU; as shown in Fig.~\ref{fig:iou}(c), when $l$ and $w$ are accurate but the orientation is slightly off, IoU drops to 0.33. If the center offset further increases, performance degrades sharply.

As shown in Fig.~\ref{fig:iou}(d), corner-aligned regression is even more sensitive to yaw errors—a minor angular deviation can reduce IoU to as low as 0.25. This indicates that corner regression is more susceptible to noise. Nevertheless, in real LiDAR point clouds, corner regions typically accumulate denser and more informative points, yielding more discriminative features. To address this trade-off, we conduct experiments on different encoding scheme and select the robust ``full-corner'' encoding scheme (Fig.~\ref{fig:iou}(e)), which fits the optimal bounding box through corner voting, thereby mitigating the adverse effects of angular errors.

\begin{figure}[ht!]
    \centering
    \includegraphics[width=0.99\linewidth]{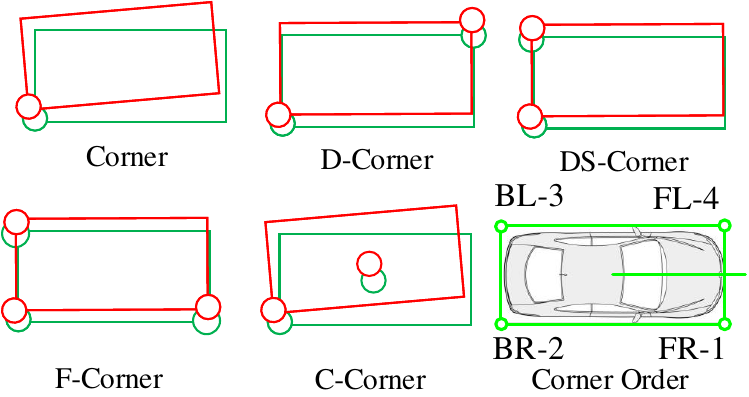}
    \caption{Five different corner encoding schemes.}
    \label{fig:corner}
\end{figure}

We reparameterize the 3D bounding box as ${B_{3D}} = \left( B_{\text{BEV}}, z_{c}, h \right)$, where $B_{\text{BEV}}$ denotes the 2D bounding box in the bird’s-eye view (BEV), and $N_{o}$ is the corner index assigned in clockwise order starting from the front-left corner, as illustrated in Fig.~\ref{fig:corner}.

\noindent\textbf{(1) Basic Corner Encoding (Corner).}  
The center of ${B_{\text{BEV}}}$ is replaced by a corner position, accompanied by a corner index identifier. The four corners are numbered clockwise as $N_{o} = 0, 1, 2, 3$, yielding ${B_{\text{BEV}}}_{\text{corner}}$:
\begin{equation}
    \begin{aligned}
        {B_{\text{BEV}}}_{\text{corner}} = \left( x_{o}, y_{o}, N_{o}, l, w, \theta \right).
    \end{aligned}
\end{equation}

\noindent\textbf{(2) Diagonal Corner Encoding (D-Corner).}  
Inspired by the corner-aware 2D image detection method proposed by Law~\textit{et al.}~\cite{law2018cornernet}, localization is achieved by predicting the positions of two diagonal corners. The orientation is then determined by the yaw angle $\theta$, eliminating the need to regress length $l$ and width $w$. This formulation is denoted as ${B_{\text{BEV}}}_{\text{d-corner}}$:
\begin{equation}
    \begin{aligned}
        {B_{\text{BEV}}}_{\text{d-corner}} = \left( x_{o1}, y_{o1}, x_{o2}, y_{o2}, N_{o}, \theta \right).
    \end{aligned}
\end{equation}
This is an anchor-free design, but it requires inferring $l$ and $w$ from the corner positions and $\theta$, and any error in $\theta$ will amplify the size estimation error.

\noindent\textbf{(3) Diagonal-Size Corner Encoding (DS-Corner).}  
To mitigate the impact of angular noise, in addition to the two diagonal corners, we explicitly predict $l$ and $w$, and then derive $\theta$ through geometric relationships. This yields ${B_{\text{BEV}}}_{\text{ds-corner}}$:
\begin{equation}
    \begin{aligned}
        {B_{\text{BEV}}}_{\text{ds-corner}} = \left( x_{o1}, y_{o1}, x_{o2}, y_{o2}, N_{o}, l, w \right).
    \end{aligned}
\end{equation}
Although this avoids error propagation from angle estimation, in LiDAR data, rear corners are often occluded due to self-shading (typically only three corners are visible), which invalidates the diagonal-corner assumption and introduces significant estimation bias.

\noindent\textbf{(4) Full Corner Encoding (F-Corner).}  
All four corner coordinates are predicted (when visible), and ${B_{\text{BEV}}}_{\text{f-corner}}$ directly encodes the object box using corner positions:
\begin{equation}
    \begin{aligned}
        {B_{\text{BEV}}}_{\text{f-corner}} = \left( x_{o1}, y_{o1}, x_{o2}, y_{o2}, x_{o3}, y_{o3}, x_{o4}, y_{o4} \right).
    \end{aligned}
\end{equation}
The yaw angle $\theta$ can be estimated from the consistency of opposite edge slopes, and edge lengths provide geometric consistency checks. Even if individual corners contain errors, their impact can be mitigated through averaging, resulting in strong robustness.

\noindent\textbf{(5) Center-Corner Encoding (C-Corner).}  
As a compromise, this scheme combines center and corner information. ${B_{\text{BEV}}}_{\text{c-corner}}$ is formulated as
\begin{equation}
    \begin{aligned}
        {B_{\text{BEV}}}_{\text{c-corner}} = \left( x_{c}, y_{c}, x_{o}, y_{o}, w \right).
    \end{aligned}
\end{equation}
It leverages one corner together with the center for localization and uses the width $w$ to constrain the geometric configuration of the box, thereby balancing the advantages of center regression and corner awareness.

\subsection{Corner Annotation Strategy}
\label{subs:corner_annotation}

In practical 3D annotation workflows, precisely locating the object center $(x_c, y_c, z_c)$ is notoriously challenging. The center often lies in empty space with sparse or no LiDAR returns, requiring annotators to mentally interpolate from visible boundaries. In contrast, corner regions, especially front-left and front-right, are typically well-supported by dense point clusters due to vehicle geometry and sensor viewing angles, making them visually salient and easier to click accurately in the BEV view. This observation motivates our shift toward corner-centric representations for robustness.
We draw two key insights: (1) navigation in the bird’s-eye view (BEV) is more intuitive than in the raw 3D point cloud\cite{meng2021towards}; and (2) annotating visible corners is easier and more accurate than estimating the box center, as corner regions are typically supported by dense point clouds, see paragraph 2, Sec.\ref{sec:intro}.

Therefore, this paper adopts a BEV-based corner annotation protocol that requires only the positions $(x_o, y_o)$ and indices $N_o$ of visible corners—no need to label all four corners. The ground height $z_g$ is estimated using the ground plane fitting method proposed by Chen~\textit{et al.}~\cite{Xiaozhi2015}.

\begin{figure*}[t]
		\vspace{-1pt}
		\centering
		\includegraphics[width=0.98\linewidth]{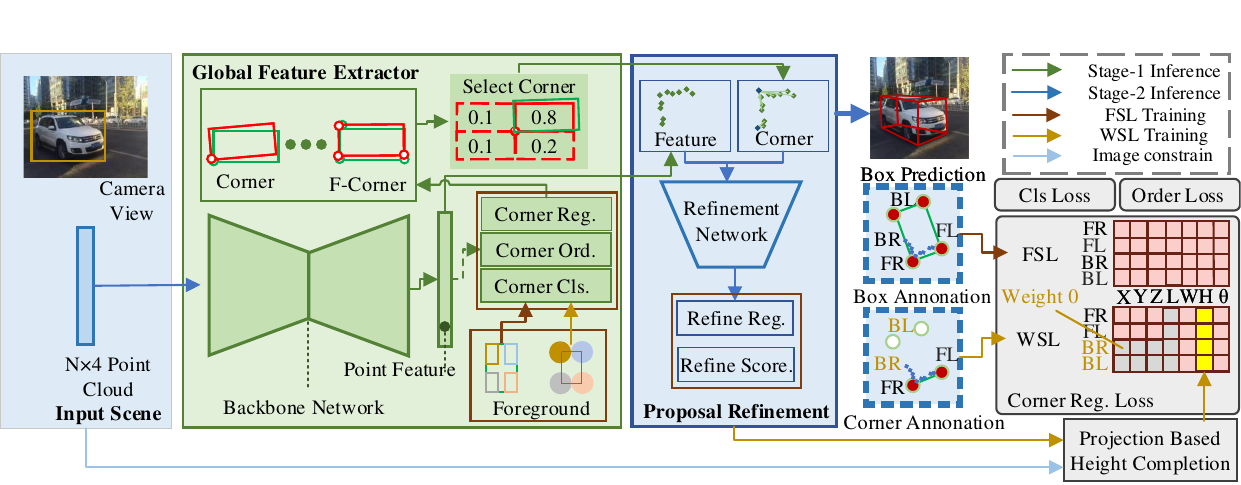}
		\vspace{-2mm}
		\caption{{The proposed corner-aware 3D object detection. While the inference route is represented with \textbf{{\color{green}green}} arrow for the first stage (stage-1) and \textbf{{\color{blue}blue}} arrow for the second stage (stage-2). The fully supervised learning part with corner-encoded 3D bounding box label is shown with \textbf{{\color{red}red arrow}}.
		The weakly supervised learning part with raw corner label and 2D geometrical constrain is shown with \textbf{{\textcolor[RGB]{191,144,0}{brown arrow}}} and \textbf{{\textcolor[RGB]{157,195,230}{shallow blue arrow}}} which will be detailed later in~Sec.\ref{subs:access_corner}.
		The framework can be trained with either full 3D bounding box annotation and weakly-supervised corner annotation, respectively.
}
{Thus, they have different \textbf{Foreground} (left for Fully Supervised Learning (FSL), right for Weakly Supervised Learning (WSL), and different loss functions.
The height information of WSL comes from geometrical constrain between corner annotation and 2D image box supervision,}
}
		\label{fig:wframework}
		\vspace{-5mm}
	\end{figure*}

We thus define the weakly supervised corner annotation as:
\begin{equation}
    \begin{aligned}
        {B_{3D}^{w}}_{\text{corner}} = \left( x_{o}, y_{o}, {z_g}_{o}, N_{o} \right).
    \end{aligned}
    \label{eq:b3d}
\end{equation}
Given any two adjacent corners (e.g., BL+FL or BR+FR), the corresponding edge length can be computed, and the yaw angle $\theta$ can be derived from the slope of the connecting line (see ``Corner Annotation'' in Fig.~\ref{fig:wframework}). Furthermore, by incorporating 2D bounding boxes from the image domain (via projection geometry constraints), the object height $h$ can be recovered (see ``Projection-based Height Prediction'' in Fig.~\ref{fig:wframework}), as detailed below.

\subsection{Corner-supervised 3D Object Detection}
\label{subs:access_corner}

We build upon PointRCNN~\cite{Shi_2019_CVPR} as the baseline framework and construct a corner-aware two-stage detector, as illustrated in Fig.~\ref{fig:wframework}.

In the first stage, a global feature extractor processes the input point cloud through PointRCNN’s encoder-decoder backbone to generate per-point features. These features are then fed into three parallel branches to predict: (i) corner offsets, (ii) corner indices, and (iii) object classification scores. To prevent boundary points from confusing the classifier, an erasure strategy is applied during foreground classification—a detail further elaborated in the experiments section.

After obtaining corner predictions, high-confidence corners are selected as region proposals. Depending on the chosen corner encoding scheme, BEV 2D box candidates are generated accordingly. For instance, under C-Corner encoding, predicted corners are clustered by index $N_o$, and for each index, the three nearest cluster centers are identified as candidate matches. The centers of lines connecting two matched corners or centroids of three matched corners are then used to infer the remaining corner clusters (with a maximum search radius of 6 meters). Finally, a least-squares fit yields the BEV box proposal. If no valid match is found, a default 6m$\times$6m region centered around the annotated corner is used instead. This results in initial BEV region proposals that crop local point clouds for the next stage.

In the second stage, the cropped point features and predicted corner estimates are concatenated and passed through a PointNet network for region proposal refinement. 

First, we train the entire network under full supervision (denoted as ``FSL'' in Fig.~\ref{fig:wframework}). Specifically, we convert the official KITTI 3D bounding box annotations into corner coordinates and use them as regression targets. Apart from replacing the center-based regression target with corner coordinates, the rest of the pipeline remains identical to the standard fully supervised framework.

Under the weakly supervised setting, the model must learn to predict a complete $B_{3D}$ from the partial signal ${B_{3D}^{w}}_{\text{corner}}$. To compensate for the missing information, we leverage geometric relationships among corners:
\begin{itemize}
    \item With only one annotated corner, the model cannot recover full geometry—it can only localize that corner.
    \item With two adjacent corners, if they lie on a long edge (e.g., FL+BL), we recover $(l, \theta)$; if on a short edge (e.g., FR+FL), we recover $(w, \theta)$.
    \item With three or more corners, both $l$, $w$, and $\theta$ become recoverable.
\end{itemize}
Denoted as ``WSL'' in Fig.~\ref{fig:wframework}, for a box annotated only with FR and FL corners, we regress the 3D positions $(x, y, z)$ of these two corners along with width $w$, and derive $\theta$ from their connecting vector. The weak-to-full learning targets are formally defined as:
\begin{equation}
    \begin{aligned}
        \text{WeaktoFull} =
        \left\{
        \begin{aligned}
            &\left(x_{o}, y_{o}, {z_g}_{o}, l, N_{o}, \theta\right) && \text{if side}, \\
            &\left(x_{o}, y_{o}, {z_g}_{o}, w, N_{o}, \theta\right) && \text{if front/rear}, \\
            &\left(x_{o}, y_{o}, {z_g}_{o}, l, w, N_{o}, \theta\right) && \text{if } \geq 3 \text{ corners}.
        \end{aligned}
        \right.
    \end{aligned}
    \label{eq:w2f}
\end{equation}

Due to the absence of complete box parameters, foreground points cannot be defined as in fully supervised methods. Instead, we define foreground points as those within 0.7 meters of any annotated corner in the BEV plane, treating them as the local neighborhood around that corner.

We then train the network for 20 epochs using ${B_{3D}^{s}}_{\text{corner}}$ as the supervision signal to learn corner regression. Unlike the fully supervised setting—which provides complete box parameters for every object—the weakly supervised approach supplies only partial annotations. To account for varying annotation density, we reweight the loss for each object $m$ by $\frac{M}{m}$ when $m \neq 0$, and assign zero weight when $m = 0$ (i.e., no corners annotated). After 20 epochs, the model acquires basic corner localization capability and can predict BEV boxes and ground height.

Next, we incorporate the manually annotated 2D bounding boxes from the KITTI dataset~\cite{KITTI} to recover the top height $z_t$ via the following geometric constraint:
\begin{equation}
    \begin{aligned}
        z_t = \min_{i \in \{1,2,3,4\}} \left\{ z_i \,\middle|\,
        K \cdot 
        \begin{bmatrix}
            u_t \\ v_i \\ 1
        \end{bmatrix}
        = P^{3\times4} \cdot
        \begin{bmatrix}
            x_i \\ y_i \\ z_i \\ 1
        \end{bmatrix}
        \right\},
    \end{aligned}
    \label{eq:constrain}
\end{equation}
where $(x_i, y_i, z_i)$ are the 3D coordinates of the four BEV corners, $u_t$ is the row coordinate of the top edge of the 2D box, $v_i$ is the column coordinate of the projected corner, $P^{3\times4}$ is the camera projection matrix, and $K$ is a scale factor. Given $(x_i, y_i)$, $P^{3\times4}$, and $u_t$, this equation can be solved to obtain four candidate $z_i$ values.

From these, we select the minimum $z_i$ as the top height $z_t$. The object height is then computed as $h = z_t - z_p$, where $z_p$ denotes the ground height. This completes the pseudo-label ${B_{3D}^{s}}_{\text{corner}} \rightarrow B_{3D}$. Using these completed labels, we continue training the model for an additional 60 epochs to learn height prediction.

\section{Experiments}\label{sec:exp}

This section presents experiments designed to evaluate the effectiveness of the proposed corner-aware learning framework. For details on the datasets and evaluation metrics. Firstly, we compare our approach against a retrained baseline model, PointRCNN~\cite{Shi_2019_CVPR}, to validate the accuracy gains brought by corner encoding and corner feature assistance. Subsequently, through ablation studies, we analyze the impact of different corner encoding schemes, foreground point definition strategies, and feature fusion mechanisms on the detector's performance, providing insights into our design choices. Finally, we assess the model's capability to leverage limited annotation information under weakly supervised settings, demonstrating its potential application in scenarios with constrained labeling costs.

    \subsection{Implementation Details}

We integrate the corner-aware detection head into the PointRCNN framework, adapting its first-stage classification and regression branches accordingly. The optimizer used is Adam, coupled with a single-cycle learning rate schedule starting at an initial learning rate of 0.001. The fully supervised model is trained for 80 epochs, while the weakly supervised model follows a two-phase training strategy: the first 20 epochs are dedicated to pre-training using only corner position supervision, followed by 60 epochs of end-to-end fine-tuning incorporating height recovery labels. All experiments were conducted on four Tesla V100 GPUs with a batch size of 16.

	\begin{table*}[t]
		\caption{\textbf{Evaluation results (\S\ref{sec:exp-improve}) on \texttt{val} set of KITTI} 3D and BEV object detection benchmark (\textbf{Car}).
AP with the threshold of $IoU$=0.7 is reported. Our result is marked with \textbf{bold font}. }
		\centering
		\vspace*{-4pt}
		\label{table:KITTIval}
		\resizebox{0.99\textwidth}{!}{
			\setlength\tabcolsep{8.5pt}
			\renewcommand\arraystretch{1.0}
			\begin{tabular}{r||c|ccc|ccc}
				\hline\thickhline
				\rowcolor{mygray}
				\rule{0pt}{10pt}
				&  	 & \multicolumn{3}{c|}{BEV@0.7} &  \multicolumn{3}{c}{3D bounding box@0.7} \\ \cline{3-8}
				\rowcolor{mygray}
				\multirow{-2}{*}{Detector~~~~}	& \multirow{-2}{*}{Modality}&Easy &Moderate &Hard &Easy &\textbf{Moderate} &Hard \\  \hline \hline \rule{0pt}{10pt}
				MV3D~\!\cite{Chen_2017_CVPR} &LiDAR+Mono &86.55 &78.10 &76.67 &71.29 &62.68 &56.56\\
				F-PointNet~\!\cite{Qi_2018_CVPR}&LiDAR+Mono &88.16&84.02&76.44&83.76&70.92&63.65\\
				MMF~\!\cite{Liang_2019_CVPR}  &LiDAR+Mono&96.66&88.16&79.60&87.90&77.86&75.57\\
				\cline{2-8} \rule{0pt}{10pt}
				VeloFCN~\!\cite{Li2016VehicleDF} &LiDAR&40.14 &32.08 &30.47  &15.20 &13.66 &15.98\\
				PIXOR~\!\cite{Yang_2018_CVPR}&LiDAR&86.79&80.75&76.60&-&-&-\\
				VoxelNet~\!\cite{Zhou_2018_CVPR} &LiDAR &89.60 &84.81 &78.57&81.97 &65.46 &62.85\\
				SECOND~\!\cite{Yan2018}&LiDAR &89.96&87.07&79.66&87.43&76.48&69.10\\
				PointPillars~\!\cite{Lang_2019_CVPR}&LiDAR &89.64&86.46&84.22&85.31&76.07&69.76\\
				Fast PointR-CNN~\!\cite{Chen_2019_ICCV}&LiDAR &90.12 &88.10 &86.24&89.12 &79.00 &77.48\\
				STD~\!\cite{Yang_2019_ICCV}&LiDAR &94.74&89.19&86.42&87.95&79.71&	75.09\\
				PointRCNN~\!\cite{Shi_2019_CVPR} &LiDAR &90.21&87.89&85.51&89.19&78.85&77.91\\			
				\hline
				\textbf{PointRCNN+Corner} &LiDAR &\textbf{93.15} &\textbf{89.43} &\textbf{88.89} &\textbf{91.09} &\textbf{82.22} &\textbf{78.05}\\
				\hline	
			\end{tabular}
		}
\vspace{-3mm}
	\end{table*}

	\begin{table*}[t]
		\caption{\textbf{Ablation study (\S\ref{subs:exp-abl}) on  \texttt{val} set of KITTI} 3D and BEV object detection benchmark (\textbf{Car}).
AP with the threshold of $IoU$=0.7 is reported. The lines with \textbf{bold font} are used for following experiments.}
		\centering
		\label{table:abl}
		\resizebox{0.99\textwidth}{!}{
			\setlength\tabcolsep{8.5pt}
			\renewcommand\arraystretch{1.0}
			\begin{tabular}{c|c|c||ccc|ccc}
				\hline\thickhline
				\rowcolor{mygray}
				\rule{0pt}{10pt}
				&   	 & &\multicolumn{3}{c|}{BEV@0.7} &  \multicolumn{3}{c}{3D bounding box@0.7} \\ \cline{4-9}
				\rowcolor{mygray}
				\multirow{-2}{*}{Method~~~~}	&\multirow{-2}{*}{Encoding~~~~}	&\multirow{-2}{*}{Earsing Region~~~~} &Easy &Moderate &Hard &Easy &\textbf{Moderate} &Hard \\  \hline \hline \rule{0pt}{10pt}
				{baseline~\!\cite{shi2020pv}}
				&center&- &90.21&87.89&85.51&89.19&78.85&77.91
				\\
				\hline \rule{0pt}{10pt}
				\multirow{5}{*}{+stage-1 promotions}
				&corner&- &92.39 &81.94 &81.73&86.81 &75.14 &72.98 \\
				&corner &small box &92.40&83.70&81.72&86.97&75.35&73.21\\
                &corner&cross region  &92.47 &88.87&86.47&87.23 &78.28 &74.29\\
				&c-corner&cross region  &92.96&84.34&84.06&88.77&75.43&73.11\\
				&d-corner&cross region &92.37 &85.96 &83.93 &88.26 &77.11 &74.87
				\\
				&ds-corner&cross region  & 92.36 &83.97 &83.72 &87.82 &75.19 &72.91
				\\
				&cc-corner&cross region  &84.83&69.89&67.65&89.97&79.18&77.01\\
				&\textbf{f-corner}&\textbf{cross region}   &\textbf{93.07}&\textbf{89.22}&\textbf{87.01} &\textbf{90.45}&\textbf{80.73}&\textbf{78.01}
				\\
				\hline \rule{0pt}{10pt}
				\multirow{2}{*}{+stage-2 promotions}
				&\multicolumn{2}{c||}{w/o corner feature}  &93.13&89.17&86.74&90.76 &80.02&77.33
				\\
				&\multicolumn{2}{c||}{\textbf{with corner feature}} &\textbf{93.15} &\textbf{89.43} &\textbf{88.89} &\textbf{91.09} &\textbf{82.22} &\textbf{78.05}\\
				\hline	
				\multicolumn{3}{c||}{learning with weak corner annotation} &90.49 &77.22 &70.25 &83.53 &65.68 &60.85\\
				\hline
			\end{tabular}
		}
		\vspace{-4mm}
	\end{table*}
	\subsection{Main Results Analysis} \label{sec:exp-improve}

Table~\ref{table:KITTIval} presents the performance comparison between our method and the retrained PointRCNN baseline on the KITTI validation set. It can be observed that, after incorporating corner-aware learning, the model achieves consistent improvements in both BEV and 3D detection tasks. Specifically, the 3D detection AP under the moderate difficulty setting increases from 78.85 to 82.22, yielding an absolute gain of 3.37. This validates the effectiveness of the proposed corner encoding and feature-assisted mechanism.

\subsection{Ablation Studies} \label{subs:exp-abl}

To further analyze the contribution of individual components, we conduct a series of ablation experiments, with results summarized in Table~\ref{table:abl}. All ablation studies are evaluated on the KITTI validation set following the official evaluation protocol, and all models are trained on the training set using the same training strategy.

\noindent\textbf{Impact of Foreground Point Definition.}  
Directly introducing corner regression into the first stage (using the ``corner'' encoding) leads to performance degradation, particularly in 3D detection (Moderate: 75.14). We attribute this to classification ambiguity caused by boundary points that simultaneously belong to multiple corner categories. To mitigate this issue, we propose an \textit{erasure strategy}: regions near box boundaries that are prone to confusion are marked as ignored, and only interior points sufficiently far from edges are retained as foreground.

\begin{figure}[t]
		\centering
		\includegraphics[width=0.98\linewidth]{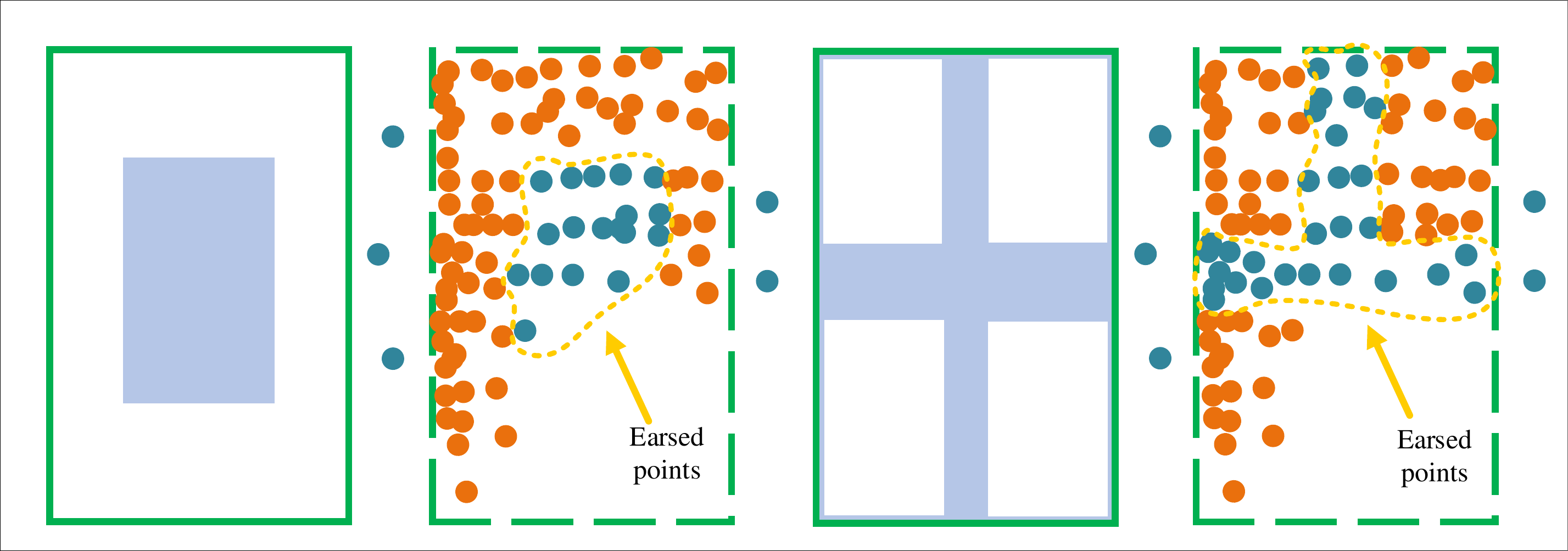}
		\caption{{The example of two erasing region strategies. The small box region strategy is on the left, and the cross region strategy is on the right.
	    The region regarded as background is colored with \textbf{{\textcolor[RGB]{157,195,230}{shallow blue}}}.
	    We mark the points that locate inside ground-truth boxes and outside the erasing region as foreground, which are colored \textbf{{\textcolor[RGB]{234,112,14}{orange}}}, and the rest of points inside box are earsed points and are colored \textbf{{\textcolor[RGB]{49,133,156}{blue}}} with background points .
		} 
		The left one is small box region, and the right one is cross region.
		}
		\label{fig:ignore}
		\vspace{-5mm}
	\end{figure}

As illustrated in Fig.~\ref{fig:ignore}, we explore two erasure designs: \textit{center-region erasure} and \textit{four-corner erasure}. Experimental results show that the former yields only marginal improvement, whereas the latter—by preserving discriminative strip-like regions along the length and width directions—better maintains informative features and improves accuracy (3D Moderate AP increases to 78.28), as reported in Table~\ref{table:abl}.

\noindent\textbf{Impact of Corner Encoding Schemes.}  
We evaluate multiple corner encoding strategies. Experimental results show that both ``c-corner'' (center + corner) and ``d-corner'' (diagonal corners) outperform the basic ``corner'' encoding, yet they still fail to fully exploit the potential of corner information. The proposed ``f-corner'' (full-corner encoding) achieves the best performance among all variants, attaining a 3D Moderate AP of 80.73—nearly 2 percentage points higher than the baseline (see Table~\ref{table:abl}). This confirms its robustness against pose perturbations.

\noindent\textbf{Role of Corner Feature Fusion in the Second Stage.}  
In the second stage, we concatenate the high-confidence corner positions predicted by the first stage as additional geometric features with the RoI point cloud and feed them into PointNet for refinement. As shown in Table~\ref{table:abl}, this strategy further improves the 3D Moderate AP from 80.02 to 82.22, demonstrating that explicit corner location cues help model object geometry more accurately.

\subsection{Effectiveness under Weak Supervision}
To simulate realistic annotation scenarios, we convert the full 3D bounding box labels in the training set into corner-level annotations: in the BEV view, if no LiDAR point exists within a 0.3-meter radius around a corner, that corner is deemed invisible and its annotation is removed. This mimics real-world occlusion-induced corner missing during manual labeling. Height information is recovered using the 2D bounding boxes provided by KITTI together with geometric projection constraints, thereby establishing a weakly supervised learning setting.

The last row of Table~\ref{table:abl} reports performance under this corner-only weak supervision. Despite the absence of complete 3D bounding box labels, the model achieves approximately 83\% (77.22 / 78.85) of the baseline’s BEV performance and 83\% (65.68 / 78.85) of its 3D performance, indicating that the proposed framework can effectively leverage sparse annotations while significantly reducing labeling cost without severe performance degradation.

	\section{Conclusion}
	In this paper, we present a weakly supervised 3D object detection method based on BEV corner-click annotation and systematically compares the impact of various corner encoding schemes on model training. Experiments on the KITTI dataset demonstrate that the ``full-corner'' encoding yields the largest performance gain among all evaluated strategies. The proposed weakly supervised framework, built upon this encoding, effectively handles the extreme case where no 3D bounding box annotations are available. In contrast to conventional approaches that rely on full 3D bounding box labels, our method jointly exploits geometric constraints and image-derived height cues to make efficient use of minimal annotation signals, offering a practical and effective pathway toward more scalable and cost-efficient 3D object detection pipelines.

	{\small
		\bibliographystyle{IEEEtran}
        \bibliography{egbib.bib}

@STRING{CVPR="Proc. IEEE Conf. Comput. Vis. Pattern Recognit."}

@STRING{ECCV="Proc. Eur. Conf. Comput. Vis."}

@STRING{ICCV="Proc. IEEE Int. Conf. Comput. Vis."}

@STRING{NIPS="Proc. Advances Neural Inf. Process. Syst."}

@STRING{TPAMI="IEEE Trans. Pattern Anal. Mach. Intell."}

@STRING{TNNLS="IEEE Trans. Neural Network and Learning Systems"}

@STRING{IV="IEEE Intelligent Vehicles Symposium"}

@INPROCEEDINGS{KITTI,
	author={Andreas Geiger and Philip Lenz and Raquel Urtasun},
	booktitle=CVPR,
	title={Are we ready for autonomous driving? {}The {KITTI} vision benchmark suite},
	year={2012},
	pages={3354--3361},
}

@InProceedings{Shi_2019_CVPR,
	author = {Shi, Shaoshuai and Wang, Xiaogang and Li, Hongsheng},
	title = {{PointRCNN}: {3D} Object Proposal Generation and Detection From Point Cloud},
	booktitle = CVPR,
	year = {2019},
	pages={770--779},
}

@InProceedings{Chen_2019_ICCV,
	author = {Chen, Yilun and Liu, Shu and Shen, Xiaoyong and Jia, Jiaya},
	title = {Fast Point {R-CNN}},
	booktitle = ICCV,
	year = {2019},
	pages={9774--9783},
}

@InProceedings{Lang_2019_CVPR,
	author = {Lang, Alex H. and Vora, Sourabh and Caesar, Holger and Zhou, Lubing and Yang, Jiong and Beijbom, Oscar},
	title = {PointPillars: Fast Encoders for Object Detection From Point Clouds},
	booktitle = CVPR,
	year = {2019},
	pages={12697--12705},
}

@INPROCEEDINGS{Xiaozhi2015,
	author={Chen, Xiaozhi and  Kaustav Kundu and  Yukun Zhu and  Andrew G. Berneshawi and  Huimin Ma and  Sanja Fidler and Raquel Urtasun},
	booktitle=NIPS,
	title={{3D} object proposals for accurate object class detection},
	year={2015},
	pages = {424--432},
}

@InProceedings{Yang_2019_ICCV,
	author = {Yang, Zetong and Sun, Yanan and Liu, Shu and Shen, Xiaoyong and Jia, Jiaya},
	title = {{STD}: Sparse-to-Dense {3D} Object Detector for Point Cloud},
	booktitle = ICCV,
	year = {2019},
	pages={1951--1960},
}

@InProceedings{Qi_2017_CVPR,
	author = {Qi, Charles R. and Su, Hao and Mo, Kaichun and Guibas, Leonidas J.},
	title = {PointNet: Deep Learning on Point Sets for {3D} Classification and Segmentation},
	booktitle = CVPR,
	year = {2017},
	pages={652--660},
}

@article{Shi_2019_part,
	title={{From Points to Parts}: {3D} Object Detection from Point Cloud with Part-aware and Part-aggregation Network},
	author={Shi, Shaoshuai and Wang, Zhe and Shi, Jianping  and Wang, Xiaogang and Li, Hongsheng},
	journal=TPAMI,
	year={2020},
}

@InProceedings{Li_2019_CVPR,
	author = {Li, Buyu and Ouyang, Wanli and Sheng, Lu and Zeng, Xingyu and Wang, Xiaogang},
	title = {{GS3D}: An Efficient {3D} Object Detection Framework for Autonomous Driving},
	booktitle = CVPR,
	year = {2019},
	pages={1019--1028},
}

@inproceedings{lee2018leveraging,
	title={Leveraging Pre-Trained {3D} Object Detection Models For Fast Ground Truth Generation},
	author={Lee, Jungwook and Walsh, Sean and Harakeh, Ali and Waslander, Steven L},
	booktitle={The IEEE Intelligent Transportation Systems Conference},
	year={2018},
	pages={2504--2510},
}

@inproceedings{zakharov2019autolabeling,
	title={Autolabeling {3D} Objects with Differentiable Rendering of {SDF} Shape Priors},
	author={Zakharov, Sergey and Kehl, Wadim and Bhargava, Arjun and Gaidon, Adrien},
	booktitle=CVPR,
	year={2019},
	pages={12224--12233},
}

@InProceedings{Li2016VehicleDF,
	title={Vehicle Detection from {3D} {Lidar} Using Fully Convolutional Network},
	author={Bo Li and Tianlei Zhang and Tian Xia},
	booktitle = {Robotics: Science and Systems},
	year={2016},
	pages={1513--1518},
}

@InProceedings{Chen_2016_CVPR,
	author = {Chen, Xiaozhi and Kundu, Kaustav and Zhang, Ziyu and Ma, Huimin and Fidler, Sanja and Urtasun, Raquel},
	title = {Monocular {3D} Object Detection for Autonomous Driving},
	booktitle = CVPR,
	year = {2016},
	pages={2147--2156},
}

@InProceedings{Yang_2018_CVPR,
	author = {Yang, Bin and Luo, Wenjie and Urtasun, Raquel},
	title = {{PIXOR}: Real-Time {3D} Object Detection From Point Clouds},
	booktitle = CVPR,
	year = {2018},
	pages={7652--7660},
}

@InProceedings{Qi_2018_CVPR,
	author = {Qi, Charles R. and Liu, Wei and Wu, Chenxia and Su, Hao and Guibas, Leonidas J.},
	title = {Frustum PointNets for {3D} Object Detection From {RGB-D} Data},
	booktitle = CVPR,
	year = {2018},
	pages={918--927},
}

@InProceedings{Liang_2019_CVPR,
	author = {Liang, Ming and Yang, Bin and Chen, Yun and Hu, Rui and Urtasun, Raquel},
	title = {Multi-Task Multi-Sensor Fusion for {3D} Object Detection},
	booktitle = CVPR,
	year = {2019},
	pages={7337--7345},
}

@InProceedings{Zhou_2018_CVPR,
	author = {Zhou, Yin and Tuzel, Oncel},
	title = {VoxelNet: End-to-End Learning for Point Cloud Based {3D} Object Detection},
	booktitle = CVPR,
	year = {2018},
	pages={4490--4499},
}

@InProceedings{Chen_2017_CVPR,
	author = {Chen, Xiaozhi and Ma, Huimin and Wan, Ji and Li, Bo and Xia, Tian},
	title = {Multi-View {3D} Object Detection Network for Autonomous Driving},
	booktitle = CVPR,
	year = {2017},
	pages={1907--1915},
}

@article{Yan2018,
	title={Second: Sparsely embedded convolutional detection},
	author={Yan, Yan and Mao, Yuxing and Li, Bo},
	journal={Sensors},
	volume={18},
	number={10},
	pages={3337},
	year={2018},
}

@article{Chen_2019,
	title={Object as Hotspots: An Anchor-Free {3D} Object Detection Approach via Firing of Hotspots},
	author={Chen, Qi and Sun, Lin and Wang, Zhixin and Jia, Kui and Yuille, Alan},
	journal={arXiv preprint arXiv:1912.12791},
	year={2019}
}

@inproceedings{yang20203dssd,
	title={{3DSSD}: Point-based {3D} single stage object detector},
	author={Yang, Zetong and Sun, Yanan and Liu, Shu and Jia, Jiaya},
	booktitle=CVPR,
	pages={11040--11048},
	year={2020},
}

@inproceedings{shi2020pv,
	title={{PV-RCNN}: Point-voxel feature set abstraction for {3D} object detection},
	author={Shi, Shaoshuai and Guo, Chaoxu and Jiang, Li and Wang, Zhe and Shi, Jianping and Wang, Xiaogang and Li, Hongsheng},
	booktitle=CVPR,
	pages={10529--10538},
	year={2020}
}

@article{bhattacharyya2021self,
	title={Self-Attention Based Context-Aware {3D} Object Detection},
	author={Bhattacharyya, Prarthana and Huang, Chengjie and Czarnecki, Krzysztof},
	journal={arXiv preprint arXiv:2101.02672},
	year={2021}
}

@inproceedings{duan2019centernet,
	title={Centernet: Keypoint triplets for object detection},
	author={Duan, Kaiwen and Bai, Song and Xie, Lingxi and Qi, Honggang and Huang, Qingming and Tian, Qi},
	booktitle=ICCV,
	pages={6569--6578},
	year={2019}
}

@inproceedings{law2018cornernet,
	title={Cornernet: Detecting objects as paired keypoints},
	author={Law, Hei and Deng, Jia},
	booktitle=ECCV,
	pages={734--750},
	year={2018}
}

@inproceedings{ye2020improving,
	title={Improving {3D} Object Detection via Joint Attribute-oriented {3D} Loss},
	author={Ye, Zhen and Xue, Jianru and Dou, Jian and Pan, Yuxin and Fang, Jianwu and Wang, Di and Zheng, Nanning},
	booktitle={2020 IEEE Intelligent Vehicles Symposium (IV)},
	pages={951--956},
	year={2020},
	organization={IEEE}
}

@inproceedings{mousavian20173d,
	title={{3D} bounding box estimation using deep learning and geometry},
	author={Mousavian, Arsalan and Anguelov, Dragomir and Flynn, John and Kosecka, Jana},
	booktitle=CVPR,
	pages={7074--7082},
	year={2017}
}

@article{yang2021auto4d,
	title={{Auto4D}: Learning to Label {4D} Objects from Sequential Point Clouds},
	author={Yang, Bin and Bai, Min and Liang, Ming and Zeng, Wenyuan and Urtasun, Raquel},
	journal={arXiv preprint arXiv:2101.06586},
	year={2021}
}

@inproceedings{meng2020weakly,
	title={Weakly Supervised 3D Object Detection from Lidar Point Cloud},
	author={Meng, Qinghao and Wang, Wenguan and Zhou, Tianfei and Shen, Jianbing and Van Gool, Luc and Dai, Dengxin},
	booktitle=ECCV,
	pages={515--531},
	year={2020},
	organization={Springer}
}

@article{meng2021towards,
	title={Towards A Weakly Supervised Framework for 3D Point Cloud Object Detection and Annotation},
	author={Meng, Qinghao and Wang, Wenguan and Zhou, Tianfei and Shen, Jianbing and Jia, Yunde and Van Gool, Luc},
	journal=TPAMI,
	year={2021},
	publisher={IEEE}
}

@article{zhao2019object,
	title={Object detection with deep learning: A review},
	author={Zhao, Zhong-Qiu and Zheng, Peng and Xu, Shou-tao and Wu, Xindong},
	journal=TNNLS,
	volume={30},
	number={11},
	pages={3212--3232},
	year={2019},
	publisher={IEEE}
}
	}
	
\end{document}